\pdfoutput=1

\documentclass[11pt]{article}

\usepackage[preprint]{acl}

\usepackage{times}
\usepackage{latexsym}

\usepackage[T1]{fontenc}

\usepackage[utf8]{inputenc}
\usepackage{microtype}

\usepackage{inconsolata}

\usepackage{graphicx}
\usepackage{booktabs}
\usepackage{multirow}
\usepackage{tabularx}
\usepackage{makecell}
\usepackage{amsmath}
\usepackage{cleveref}

\usepackage{graphicx}
\usepackage{subcaption}
\usepackage{algorithm}
\usepackage{algorithmic}
\usepackage{supertabular}
\title{What is in a name? Mitigating Name Bias in Text Embeddings via Anonymization}

\author{Sahil Manchanda  \and Pannaga Shivaswamy \\
        Pocket FM, India \\ { \{sahil.manchanda, pannaga.s\}@pocketfm.com}}

\begin{document}
\maketitle
\begin{abstract}
    Text-embedding models often exhibit biases arising from the data on which they are trained. In this paper, we examine a hitherto unexplored bias in text-embeddings: bias arising from the  presence of \textit{names} such as persons, locations, organizations etc. in the text. Our study shows how the presence of \textit{name-bias} in text-embedding models can potentially lead to erroneous conclusions in  assessment of thematic similarity. {\em Text-embeddings can mistakenly indicate similarity between texts based on names in the text, even when their actual semantic content has no similarity or indicate dissimilarity simply because of the names in the text even when the texts match semantically}. We first demonstrate the presence of name bias in different text-embedding models and then propose \textit{text-anonymization} during inference which involves removing references to names, while preserving the core theme of the text. The efficacy of the anonymization approach is demonstrated on two downstream NLP tasks, achieving significant performance gains. Our simple and training-optimization-free approach offers a practical and easily implementable solution to mitigate name bias.

\end{abstract}

\section{Introduction}
Text-embedding models, which convert raw text such as sentences/paragraphs into concise numerical representations, have become fundamental tools for downstream NLP tasks in fields such as healthcare, education, law and scientific research~\citep{chrysostomou2022flexible, reimers2019sentence, tenney2019bert, nie2024text, sun2019bert4rec}.
A cosine similarity between embeddings is typically used \cite{zhang2019bertscore, mathur2019putting} although other types of similarities  \cite{steck2024cosine} are also possible. With a similarity measure, the goal is to find which two texts are similar to or different from one another.  For simplicity, we will use \textit{text-embedding model} to refer to models that convert text to an embedding.

Many text-embedding models are often trained on large amounts of Internet text. This data can inadvertently contain biases of various kinds, reflecting social prejudices and stereotypes. As a result, these models can generate biased embeddings, reinforcing harmful stereotypes or discriminating against certain cultural groups, genders, etc.~\citep{gallegos2024bias, li2023survey, rakivnenko2024bias}. Furthermore, the presence of bias in models could lead to embeddings that disproportionately emphasize particular parts of the text, consequently failing to capture the true semantics and themes within the text~\citep{rakivnenko2024bias}.

While important, existing studies on biases, predominantly examine biases in text-embedding models mostly related to gender, geography, race, religion etc.~\citep{rakivnenko2024bias, may2019measuring, bolukbasi2016man, kotek2023gender, nghiem2024you}. In this paper, we demonstrate that text-embedding models exhibit significant bias towards \textit{names} within the text. To illustrate this, we begin with a motivating example in Table~\ref{tab:opening_example_story}. We present a simple narrative (\textit{Story 1}). We then show a similar plot while substituting the name of the main character in (\textit{Story 2}). In the third narrative (\textit{Story 3}), we introduce a distinct and contradicting storyline from \textit{Story 1 }while retaining the original character names. We embed all three stories using text-embedding models.  We observe that the similarity between Story 1 and Story 3, despite their differing plots, is consistently higher than the similarity between Story 1 and Story 2, which share highly semantically similar plots but differ in character names. This is very counter-intuitive since the text-embedding models seem to prioritize name similarity over the text's narrative structure. While this is admittedly an illustrative example, we proceed to generate numerous such narratives and conduct a thorough investigation of this bias in our experiments. {\em We emphasize here that our study investigates thematic and semantic similarities within textual data while acknowledging certain applications involving text tied to specific individuals or locations, our primary focus lies on the broader thematic context rather than characters in the text.}

Our observation reveals a critical issue that can significantly impact applications that rely on semantic similarity, including semantic search, information retrieval, and plagiarism detection~\citep{minaee2024large, pudasaini2024survey}: consider the challenge of accurately assessing the similarity between two stories/plots with identical underlying meanings but distinct character names. Current methods may erroneously classify these stories as dissimilar, leading to inconsistent and unreliable results. Further, based upon our investigation, we would like to mention upfront that the issue is not confined to certain cultures, cross-culture, but is universal in the sense that the name bias issue occurs in a very broad sense.

\begin{table}[ht!]
\centering
\centering
\scalebox{0.8}{
\begin{tabular}{lp{7.6cm}}

 \hline

Story Id & Text \\

\hline

Story 1 & \textit{Alejandro} gently examined the injured bird. He gave it food. \\


Story 2 & \textit{Jelani} tenderly inspected the wounded bird and gave it a meal to eat.\\


Story 3 & \textit{Alejandro} tracked the injured bird. He used it as his food. \\


\end{tabular} 
}

\vspace{1em} 

\scalebox{0.61}{
\begin{tabular}{lcc}
\toprule
Model & \multicolumn{2}{c}{Cosine Similarity} \\
\cmidrule(lr){2-3}
& Story1, Story 2 $\uparrow$ & Story 1, Story 3 $\downarrow$\\
\midrule

                    all-mpnet-base-v2 &                                         0.755 &                                0.778 \\
                 all-distilroberta-v1 &                                         0.780 &                                0.798 \\
                     all-MiniLM-L6-v2 &                                         0.660 &                                0.853 \\
                               gemini &                                         0.864 &                                0.848 \\
           multi-qa-distilbert-cos-v1 &                                         0.579 &                                0.907 \\
              paraphrase-MiniLM-L6-v2 &                                         0.775 &                                0.855 \\
 distiluse-base-multilingual-cased-v1 &                                         0.752 &                                0.889 \\
 distiluse-base-multilingual-cased-v2 &                                         0.742 &                                0.875 \\
paraphrase-multilingual-MiniLM-L12-v2 &                                         0.836 &                                0.840 \\
            msmarco-distilbert-cos-v5 &                                         0.584 &                                0.817 \\
           multi-qa-mpnet-base-cos-v1 &                                         0.694 &                                0.854 \\
                        voyage-3-lite &                                         0.780 &                                0.868 \\
               text-embedding-3-small &                                         0.755 &                                0.826 \\
               text-embedding-3-large &                                         0.741 &                                0.808 \\

\hline
\end{tabular}
}
\caption{\textbf{Impact of names on similarity}: 
We see that Story 1 is similar to Story 2 but has different person names(\textit{Alejandro}, \textit{Jelani}). Story 3 is different from Story 1 but has same name (\textit{Alejandro}) as Story 1. We observe that, in most embedding models a different story with opposite meaning and same name(\textit{Alejandro}) is getting a higher similarity score in comparison to the same story with different names. \label{tab:opening_example_story}}
\vskip -0.1in
\end{table}

Having briefly revealed the issue of name bias in text-embedding models, we  outline our contributions in the work:

    \looseness=-1
    First, we identify bias arising from names in textual content. Although several forms of biases have been studied in the past (see Sec.\ref{sec:related}), to the best of our knowledge, our work is the first that specifically looks at bias associated with names and how they can influence the embeddings coming out of embedding models. Toward this end, we propose a benchmarking study to comprehensively assess this bias.
    
    \looseness=-1
    Second, we propose a simple \textit{inference-time text-anonymization} technique designed to overcome the identified bias. Our method does not require any model fine-tuning or retraining of the text-embedding models. The approach offers a simple, intuitive, and effective way to mitigate the problem rather than relying on complex computations.

    \looseness=-1
    Third, we conducted extensive experiments to study the identified problem in detail on a variety of text-embedding models and tasks. Our results demonstrate that our anonymization approach effectively reduces name bias within embeddings in semantic similarity and downstream tasks. 

    \vspace{-0.05in}

\section{Related Work}
\vspace{-0.1in}
\label{sec:related}
\paragraph{Biases in Text-embedding models:} Text-embedding models while powerful, can inadvertently reflect and amplify existing biases and prejudices; there is vast research understanding and mitigating bias in such models. For example, there is work focusing on models that investigate under-representation or misrepresentation of  specific groups, such as LGBTQ+ individuals, leading to skewed or inaccurate outcomes~\citep{may2019measuring, bolukbasi2016man, cheng2021fairfil}. Another type of study focuses on gender bias in word embeddings models~\citep{rakivnenko2024bias}.  The study highlights a concerning issue i.e many embedding models associate specific occupations with particular genders. ~\citet{nikolaev2023representation} studied biases at a sentence-level in sentence transformers influenced by different parts of speech such as common nouns, adverbs etc.
While we discuss text-embedding model, it is important to highlight works that investigate bias within Large Language Models (LLMs) for text-generation which are a part of this ecosystem~\citep{gallegos2024bias}. ~\citet{schwobel2023geographical} observed "geographical erasure" where certain regions are underrepresented in LLM outputs. ~\citet{manvi2024large}  showed that LLMs often favor developed regions and exhibit negative biases towards locations with lower socioeconomic conditions, particularly on subjective topics such as attractiveness and intelligence. Further, some works have also investigated  cross-cultural biases in LLMs for text generation~\citep{naous2023having, ramezani2023knowledge, cao2023assessing, arora2022probing}. Compared to the above work, we investigate name-bias in text-embeddings, an area not previously explored in existing research to the best of our knowledge.

\paragraph{Debiasing methods:} Various approaches have been proposed to tackle different kinds of biases in text-embedding models highlighted above. One common technique to remove such biases is to update the training dataset and make it unbiased and re-train the model~\citep{brunet2019understanding, ngo2021mitigating}. Another paradigm involves applying approaches such as disentanglement or alignment where models are fine-tuned to remove biases associated with concepts such as gender, religion etc.~\citep{kaneko2021debiasing, guo2022auto, kenneweg2024debiasing}. 
An alternative approach involves post-processing of the embeddings. Specifically, it involves adding a debiasing module after encoders to filter out certain biases in the representations~\citet{cheng2021fairfil}. For more details on this topic, we refer the reader to survey by ~\citet{li2023survey} for more details.

We  emphasize some key considerations based upon the discussions above. Firstly, all the aforementioned techniques require an optimization phase, involving either retraining the initial model, fine-tuning with a modified loss or post-processing of the generated embeddings. Secondly, these methods are often designed to address specific bias types, such as social, gender, or religious biases. Notably, the identification and mitigation of name bias has not been previously explored to our knowledge.
\section{Understanding name bias}
\label{sec:benchmarking}
\looseness=-1
In this section, we investigate the presence of bias within text-embedding models related to names. Our primary objective is to investigate the influence of names containing identity-specific information on the resulting text embeddings, while ensuring the semantic structure of the text remains unchanged.

\subsection{Benchmarking Methodology}
\looseness=-1
To understand the impact of bias associated with names, we systematically replace instances of names in text with alternatives. For the sake of simplicity, in this section, we focus on person names and country names\footnote{We also study the impact of  perturbation of person names only.}. Given a text, we first identify instances of person and country names in the text.\footnote{The datasets used for benchmarking are described in Sec.~\ref{benchmark:datasets} } To study bias w.r.t. person names, we replace each person name in the text with a randomly sampled name from a list of person names. In the text, all instances of the same person are replaced by the same sampled name. Similarly, country names are replaced with a random country name sampled from a predefined list of countries. This process only changes the person names and countries and does not change the original structure or meaning of the text.

Formally, given a universe of $n$ person names $P=\{p_1, p_2, p_3 \cdots p_n\}$, and $l$ Country names $C=\{c_1, c_2, c_3 \cdots c_l\}$, we apply algorithm~\ref{alg:perturbation_text} for a given text $T$ to obtain a perturbed text $T'$. 

\begin{algorithm}
\small
\caption{Perturb Text for Benchmarking }
\begin{algorithmic}[1]
 	 \REQUIRE   $P:$ List of Person names, $C:$ List of Country names.
     \STATE {\bfseries Input:} Text $T$ 
\STATE {\bfseries Output:} Text $T'$ with replaced entities
     \STATE Initalize: $T' \leftarrow T$ 

        \STATE{\bfseries Identify Entities:}  
    \STATE $\quad$ Identify all occurrences of person names $IP$ in $T'$.
    \STATE $\quad$ Identify all occurrences of country names $IC$ in $T'$. 

    \STATE{\bfseries Perturbation:}  
    \FOR{each identified person $ip \in IP$ in text $T'$} \label{alg_line_bench_person_repl_beg}
            \STATE Randomly select a name $p_k \in P$ without \\
            replacement.
           \STATE Replace all occurrences of $ip$ with $p_k$ in text $T'$.
    \ENDFOR \label{alg_line_bench_person_repl_end}

        \FOR{each identified country $ic \in IC$ in text $T'$}   \label{alg_line_bench_country_repl_beg}
        \STATE Randomly select a country name $c_k \in C$ without \\replacement.
        \STATE Replace all occurrences of $ic$ with $c_k$ in text $T'$.
    \ENDFOR \label{alg_line_bench_country_repl_end}
  \STATE {\bfseries Return} $T'$  \COMMENT{Perturbed Text}

\end{algorithmic}
\label{alg:perturbation_text}
\end{algorithm}

Applying  Algorithm~\ref{alg:perturbation_text} gives one perturbation $T'$ for text $T$. We generate $K{=}20$ such perturbations capturing a wider range of person and country names. The names used for replacement are present in Table ~\ref{tab:names_used_benchmark} in Appendix and we have names from many different cultures/countries. Note that the steps ~\ref{alg_line_bench_person_repl_beg}-\ref{alg_line_bench_person_repl_end} and ~\ref{alg_line_bench_country_repl_beg}-\ref{alg_line_bench_country_repl_end}   in perturbation algorithm can be done in isolation and can be applied independently based upon the use-case.  An illustrative example of a perturbation is presented in Table~\ref{tab:perturbation_example}.

\begin{table}[ht]
\centering
\small
\begin{tabular}{|p{3.5cm}|p{3.5cm}|}
\hline
Original Text($T$)  & Perturbed Text 1($T'_1$) \\
\hline
\textbf{Mike} has been living in \textbf{Belgium} for five years and made a fortune by winning a lottery. \textbf{Mike} spent most of his money on treatment of his brother \textbf{Donald} who was suffering from Lung Cancer. & \textbf{Dwayne} has been living in \textbf{France} for five years and made a fortune by winning a lottery. \textbf{Dwayne} spent most of his money on treatment of his brother \textbf{Shawn} who was suffering from Lung Cancer. \\
\hline
\end{tabular}
\caption{Example of text perturbation.}
\label{tab:perturbation_example}
\end{table}

\looseness=-1
The objective is to determine the degree of semantic divergence observed between perturbed text instances, resulting from the replacement of names and countries, by examining their embeddings. As discussed above, for a text $T$ we create $K$ perturbations  $\left\{ T'_i \mid 1 \leq i \leq K \right\}$. Each of these $K$ perturbed text versions were processed through a text-embedding model, to obtain its corresponding embedding.
Subsequently, to capture the distance between the perturbed text's embeddings with each other, we calculate the pairwise cosine similarity between all $K$ embeddings.  For example, if a text sample has $K{=}20$ perturbations, we get $\frac{K \times (K-1)}{2}= 190$ similarity scores. Given $N$ such text samples in a dataset, to arrive at a single metric, we first compute pairwise cosine similarities(between the perturbed text embeddings) for a given text, excluding the self-similarity comparisons (i.e., the similarity of a perturbed text embedding to itself). For $N$ samples, we obtain $N \times \frac{K \times (K-1)}{2} $ similarity scores. Let $emb_{si}$ refer to the embedding of $i^{th}$ perturbation of sample $s$ where $s \in \{1, 2, ..., N\}$ and $i \in \{1, 2, ..., K\}$.  Then, average similarity across $N$ samples is defined as: \vspace{-0.09in}
\begin{equation}
\nonumber
    \frac{1}{N \times \frac{K(K-1)}{2}} \sum_{s=1}^{N} \left[ \sum_{i=1}^{K} \sum_{\substack{j=1 \\ j \neq i}}^{K} {Sim}(emb_{si}, emb_{sj}) \right]
\end{equation}
\normalsize
\vspace{-0.1in}

 A higher average similarity indicates that the perturbed texts are closer to each other in the semantic space, suggesting less deviation. Conversely, a lower average similarity score suggests a higher degree of deviation from the expected semantic relationship. It suggests that the embedding model exhibits a bias towards names in the text, potentially affecting its ability to accurately capture the theme of the text.

\subsection{Candidate Text-embedding Models}
We analyzed a diverse set of leading text embedding models from academia and industry. This includes models explicitly trained on diverse languages and tasks such as semantic search, question-answering etc. We include models such as \textit{multi-qa-distilbert-cos-v1} and \textit{multi-qa-mpnet-base-cos-v5} for question answering, and \textit{paraphrase-MiniLM-L6-v2} and \textit{paraphrase-multilingual-MiniLM-L12-v2} for identifying semantic similarity~\citep{reimers2019sentence}. Other notable models include \textit{all-mpnet-base-v2, all-distilroberta-v1}, and \textit{all-MiniLM-L6-v2}, designed for general-purpose text representation~\citep{reimers2019sentence}. Additionally, multilingual models like \textit{distiluse-base-multilingual-cased-v1} and \textit{distiluse-base-multilingual-cased-v2 }are also included~\citep{reimers-2020-multilingual-sentence-bert}. We also include \textit{msmarco-distilbert-cos-v5} specialized model for search~\citep{reimers2019sentence}. Additionally, we also choose cutting-edge models which are not open-source namely \textit{text-embedding-3-small} and  \textit{text-embedding-3-large} from Open AI~\citep{OpenAI2024}, \textit{gemini} from Google~\citep{team2023gemini} and \textit{voyage-3-lite} from Voyage AI~\citep{VoyageAI2024}.

\subsection{Benchmark Datasets }
\label{benchmark:datasets}

    \textbf{CMU Movie Dataset~\citep{bamman2013learning}:} The CMU Movie dataset primarily consists of textual plot summaries of movies spanning multiple sentences. These summaries are typically short , concise descriptions of the main events and storylines within a film. They often include key characters, conflicts, and resolutions. 
    
  \textbf{CMU Book Dataset~\citep{bamman2013new}:} Similar to CMU Movie, the core of this dataset consists of concise multiple sentence summaries of books. These summaries  capture the main plot points, key characters, and themes.   



\looseness=-1
We select plots where the number of words are less than $250$ which is within token limit of most models under consideration\footnote{For each embedding model, we evaluate its performance only on samples which are within the limits of its maximum context window.}. 

\begin{table}[h]
\centering

\scalebox{0.7}{
\begin{tabular}{|ll|}
\hline
 Model Name &  \makecell{Cosine sim  \\ per perturbation pair}   \\
\toprule
    all-mpnet-base-v2 &                  0.774 $\pm$ 0.001 \\
                 all-distilroberta-v1 &                  0.768 $\pm$ 0.001 \\
                     all-MiniLM-L6-v2 &                  0.706 $\pm$ 0.001 \\
                               gemini &                    0.885 $\pm$ 0.0 \\
           multi-qa-distilbert-cos-v1 &                  0.733 $\pm$ 0.001 \\
              paraphrase-MiniLM-L6-v2 &                  0.742 $\pm$ 0.001 \\
 distiluse-base-multilingual-cased-v1 &                  0.786 $\pm$ 0.001 \\
 distiluse-base-multilingual-cased-v2 &                  0.795 $\pm$ 0.001 \\
paraphrase-multilingual-MiniLM-L12-v2 &                   0.75 $\pm$ 0.001 \\
            msmarco-distilbert-cos-v5 &                  0.681 $\pm$ 0.001 \\
           multi-qa-mpnet-base-cos-v1 &                  0.743 $\pm$ 0.001 \\
               text-embedding-3-small &                    0.742 $\pm$ 0.0 \\
               text-embedding-3-large &                    0.779 $\pm$ 0.0 \\
                        voyage-3-lite &                     0.76 $\pm$ 0.0 \\
\bottomrule

\end{tabular}}
\caption{\textbf{Bias Measurement on CMU Movie dataset}. For each show, we create \textbf{$K{=}20$ perturbations} by replacing person names and country names.  In this experiment, we used plot samples that contain both person and country names but does not mention any city/town/village/nationality keywords(Spanish, American etc.) in order to minimize the impact of other variables. \label{tab:model_biases_cmu_movies}We report the mean and the std. error rounded off to $3$ decimal places. }
\end{table}

\begin{table}[h!]
\centering

\scalebox{0.7}{
\begin{tabular}{|ll|}
\hline
 Model Name &  \makecell{Cosine sim  \\ per perturbation pair}   \\
\toprule

                    all-mpnet-base-v2 &                  0.777 $\pm$ 0.001 \\
                 all-distilroberta-v1 &                  0.778 $\pm$ 0.001 \\
                     all-MiniLM-L6-v2 &                  0.693 $\pm$ 0.001 \\
                               gemini &                     0.89 $\pm$ 0.0 \\
           multi-qa-distilbert-cos-v1 &                  0.743 $\pm$ 0.001 \\
              paraphrase-MiniLM-L6-v2 &                  0.735 $\pm$ 0.002 \\
 distiluse-base-multilingual-cased-v1 &                  0.777 $\pm$ 0.001 \\
 distiluse-base-multilingual-cased-v2 &                  0.785 $\pm$ 0.001 \\
paraphrase-multilingual-MiniLM-L12-v2 &                  0.746 $\pm$ 0.002 \\
            msmarco-distilbert-cos-v5 &                  0.707 $\pm$ 0.001 \\
           multi-qa-mpnet-base-cos-v1 &                   0.75 $\pm$ 0.001 \\
               text-embedding-3-small &                  0.761 $\pm$ 0.001 \\
               text-embedding-3-large &                  0.795 $\pm$ 0.001 \\
                        voyage-3-lite &                  0.781 $\pm$ 0.001 \\
\bottomrule
\end{tabular}}

\caption{\textbf{Bias Measurement on CMU Books dataset.} We follow the same evaluation setup as in Table ~\ref{tab:model_biases_cmu_movies}.\label{tab:model_biases_cmu_book}}
\end{table}

\subsection{Analyzing Bias}
\looseness=-1
In Table~\ref{tab:model_biases_cmu_movies} and ~\ref{tab:model_biases_cmu_book} we observe a significant deviation in the average cosine similarity which should be close to one if the cosine similarity captured the real semantic similarity rather than information in names present in the text\footnote{We also experimented by using euclidean distance instead of cosine similarity in Tab.~\ref{tab:model_biases_cmu_book_euclidean} in Appendix. The conclusion remained similar and therefore we proceeded with cosine similarity for remaining experiments.}. Any deviation from one indicates that the embeddings are heavily biased by the choice of names rather than from the similarity of the text. Models like $msmarco{-}distilbert{-}cos{-}v5$ exhibit significant sensitivity to changes in person and country names, as evidenced by an average cosine similarity $\approx0.7$. This suggests that the model's embeddings may be heavily influenced by specific entities rather than capturing the underlying semantic meaning of the text. Observations from the evaluation of both datasets suggest that $gemini$ is the least biased model among all models considered. However, we observe that even \textit{gemini's} score is still far away from one indicating room for improvement.

\looseness=-1
In the above experiment, we replaced the names of people and countries and generated a perturbed text. One may ask: how much of the bias is from country name versus person names? To study this, we considered an experiment in which we perturbed the text by only replacing person names while keeping the country names as they were in the original text. We also examined variations in which all the perturbed names are sampled from the same country and demonstrate that bias persists even if text samples differ only by person names even from the same country. These results can be found in the App.~\ref{sec:person_name_perturbation}.

\label{sec:method}

\section{Methodology: Overcoming Bias through Anonymization}
Previously, we showed that how just changing person names/country names can impact the embeddings significantly. In this section, we introduce a simple \textit{inference-time anonymization} technique to mitigate the bias caused by names. The core idea is to mitigate the influence of names on embeddings, and making the resulting \textit{debiased} anonymized embeddings to be more generalizable and less prone to biases related to particular individuals or entities.  

The anonymization of a text $T$ during inference is achieved through the following process. We first identify in $T$, occurrences of desired entities such as person names, locations and organizations relevant to the use case. We \textit{anonymize} the text by removing those occurrences from $T$.  The anonymized text referred to as $T_{anon}$ retains the overall structure and meaning of the original text $T$ while removing any specific references to person names etc.  This anonymization can be achieved via tools such as Large Language Models(LLMs)~\citep{zhao2023survey} or Named Entity Recognition tools~\citep{jehangir2023survey}. In our work, we used \textit{gemini} and \textit{anthropic.claude-3-5-sonnet} text-generative models for anonymization using prompts. Depending upon the use-case, different names in text such as person names, cities, countries, organizations can be removed. We would like to clarify that the same process of anonymization can also be done through Named-Entity Recognition(NER) tools~\citep{jehangir2023survey}, however in our initial experiments we found LLMs to be more accurate. Sample prompts for anonymization are presented in Table ~\ref{tab:prompts}. Post anonymization, the embeddings become independent of identity specific details such as person names/ country names etc.~\footnote{ The type of anonymization i.e removing person names and/or country names and/or city names etc. used determines the exact level of independence.} Overall, the \textit{debiased} embeddings generated on anonymized text promise reduced sensitivity to biases associated with particular individuals or entities. Note that the embeddings generated for sentences that differ solely in their named entities (e.g., character names) will now have a cosine similarity of 1.
\noindent
An alternate to removing \textit{named} content for anonymization is to replace names with specific non-identifying placeholder words. This approach with its associated challenges is further examined in App. ~\ref{sec:anon_vs_replacement}.


\begin{table}[h!]
\centering
\scalebox{0.8}{
\begin{tabular}{p{2cm}p{6.5cm}}
\hline
 Purpose & Prompt \\
\hline
 Remove person names and location names & Given below text, please COMPLETELY DELETE all Person/Character names which are PROPER NOUNS and City/ Country/ Village/ Town/ Continent/ River/  Organization names which are PROPER NOUNS etc. Wherever they occur replace with empty string. Completely remove them and not anything else. Do not delete monument/landmark names like Eiffel tower etc. Do not remove He/She/him/her etc.. Output contains the modified text only....  The text is provided below :::: \\ \hline
 Remove person names only & Given below text, please COMPLETELY DELETE all Person/Character names which are PROPER NOUNS. Wherever they occur replace with empty string. Completely remove them and not anything else. Do not remove He/She/him/her etc.. Output contains the modified text only....  The text is provided below :::: \\
\hline
\end{tabular}}
\caption{Prompts for Anonymization. In our experiments, we select the first prompt. Based upon the use case, the suitable prompt can be selected or modified accordingly. }
\label{tab:prompts}
\end{table}

\begin{table*}[ht!]
\centering
\scalebox{0.8}{
\begin{tabular}{lp{5cm}|p{4cm}|p{3.1cm}}
\toprule
                                
                               Model  & \textit{Original text:} The \textit{(query, positive)} paragraphs share the same meaning but different person/location names. The \textit{(query, negative)} paragraphs share different meaning but same person/location names. &  \textit{Identical Names:} The (\textit{query}, \textit{positive},  \textit{negative}) paragraphs in the same triplet contain the same person/location names. &  \textit{Anonymized text:} Anonymization applied to (\textit{query}, \textit{positive},  \textit{negative}) paragraphs. \\
                            
\midrule

                    all-mpnet-base-v2 &                   0.19 &                        0.96 &       0.98 $\pm$ 0.0071 \\
                 all-distilroberta-v1 &                   0.36 &                        0.97 &      0.975 $\pm$ 0.0106 \\
                     all-MiniLM-L6-v2 &                   0.09 &                        0.94 &       0.99 $\pm$ 0.0071 \\
                               gemini &                   0.71 &                        1.00 &           1.0 $\pm$ 0.0 \\
           multi-qa-distilbert-cos-v1 &                   0.07 &                        0.97 &       0.97 $\pm$ 0.0071 \\
              paraphrase-MiniLM-L6-v2 &                   0.14 &                        0.98 &          0.98 $\pm$ 0.0 \\
 distiluse-base-multilingual-cased-v1 &                   0.27 &                        0.95 &          0.94 $\pm$ 0.0 \\
 distiluse-base-multilingual-cased-v2 &                   0.26 &                        0.98 &          0.96 $\pm$ 0.0 \\
paraphrase-multilingual-MiniLM-L12-v2 &                   0.21 &                        1.00 &          0.99 $\pm$ 0.0 \\
            msmarco-distilbert-cos-v5 &                   0.10 &                        0.92 &      0.955 $\pm$ 0.0035 \\
           multi-qa-mpnet-base-cos-v1 &                   0.08 &                        0.97 &           1.0 $\pm$ 0.0 \\
               text-embedding-3-small &                   0.12 &                        1.00 &           1.0 $\pm$ 0.0 \\
               text-embedding-3-large &                   0.21 &                        1.00 &           1.0 $\pm$ 0.0 \\
                        voyage-3-lite &                   0.18 &                        1.00 &           1.0 $\pm$ 0.0 \\

\bottomrule
\end{tabular}}
\caption{\textbf{Evaluation on Task 1: Semantic Similarity Task.} AUC scores obtained on Semantic Similarity Task. Our proposed strategy of anonymization achieves high quality results across all models. Mean and standard error are reported based on results from two separate LLM runs for anonymization. }\label{tab:sts_generated_with_reference_metric}
\end{table*}


\begin{table*}[h!]
    \centering
    \scalebox{0.9}{
    \begin{tabular}{|c|p{5cm}|p{7cm}|p{1cm}|p{1cm}|}
    \toprule
     & Query & Pos/Neg & Sim score & Label \\
    \hline
    \multirow{2}{*}{Original} & \multirow{2}{5cm}{Alejandro quickly ran to the store to buy a cold drink. He was eager to have a glass of cold drink.} & \textcolor{blue}{\textbf{POS:} Quickly, Hiroki dashed to the local market to procure some cold drinks. He was yearning for a chilled glass of cold drink.} & 0.58 & 1 \\
    & & \textcolor{red}{\textbf{NEG:} Alejandro has stopped buying cold drinks from market. He only drinks cold drinks made at home.} & 0.69 & 0 \\
    \cline{3-5}
    \multirow{2}{*}{Anonymized} & \multirow{2}{5cm}{quickly ran to the store to buy a cold drink. He was eager to have a glass of cold drink.} & \textcolor{blue}{\textbf{POS:} Quickly,  dashed to the local market to procure some cold drinks. He was yearning for a chilled glass of cold drink.} & 0.80 & 1 \\
    & & \textcolor{red}{\textbf{NEG:} has stopped buying cold drinks from market. He only drinks cold drinks made at home.} & 0.47 & 0 \\
    \hline
    \end{tabular}}
     \centering
     \scalebox{0.9}{
    \begin{tabular}{|c|p{5cm}|p{7cm}|p{1cm}|p{1cm}|}
    \hline
    \multirow{2}{*}{Original} & \multirow{2}{5cm}{Ganga and Yamuna are two mighty rivers. They are lifelines for millions of people in the region.} & \textcolor{blue}{\textbf{POS:} Yangtze is a mighty river. It is a long river and is the lifeline for millions of people in the region.} & 0.54 & 1 \\
    & & \textcolor{red}{\textbf{NEG:} Ganga and Yamuna are two sisters. They had their schooling in the region and schooling provided a lifeline for them.} & 0.70 & 0 \\
    \cline{3-5}
    \multirow{2}{*}{Anonymized} & \multirow{2}{5cm}{and are two mighty rivers. They are lifelines for millions of people in the region.} & \textcolor{blue}{\textbf{POS:} is a mighty river. It is a long river and is the lifeline for millions of people in the region.} & 0.70 & 1 \\
    & & \textcolor{red}{\textbf{NEG: }and are two sisters. They had their schooling in the region and schooling provided a lifeline for them.} & 0.56 & 0 \\
    \hline
    \end{tabular}}
    \caption{Examples showing impact of anonymization on semantic similarity using embeddings created by \textit{msmarco-distilbert-cos-v5}.}
    \label{tab:sim_change_deloc_msmarco}
    \end{table*}

\begin{table*}[ht!]
\centering
\scalebox{0.75}{

\begin{tabular}{lll|ll}
\toprule
                                model &  \makecell{Spearman-correlation \\ 
                                (Original Text)} & \makecell{Spearman-correlation \\ (Anonymized)} & \makecell{Pearson-correlation \\ (Original Text)} & \makecell{Pearson-correlation \\ (Anonymized)}  \\
\midrule

                    all-mpnet-base-v2 &                              0.262 &                 \textbf{0.344  $\pm$ 0.001} &                              0.321 &                \textbf{0.364  $\pm$ 0.002} \\
                 all-distilroberta-v1 &                              0.245 &              \textbf{0.327  $\pm$ 0.007} &                              0.302 &                \textbf{0.37  $\pm$ 0.003} \\
                     all-MiniLM-L6-v2 &                              0.251 &                 \textbf{0.33  $\pm$ 0.003} &                              0.282 &                \textbf{0.354  $\pm$ 0.006} \\
                               gemini &                              0.381 &                \textbf{0.39  $\pm$ 0.001} &                              \textbf{0.456} &                0.436  $\pm$ 0.003 \\
           multi-qa-distilbert-cos-v1 &                              0.240 &                 \textbf{0.292  $\pm$ 0.002} &                              0.269 &                \textbf{0.316  $\pm$ 0.007} \\
              paraphrase-MiniLM-L6-v2 &                              0.283 &                \textbf{0.352  $\pm$ 0.005 }&                              0.317 &                  \textbf{0.37  $\pm$ 0.0} \\
 distiluse-base-multilingual-cased-v1 &                              0.282 &                \textbf{0.356  $\pm$ 0.001} &                              0.325 &               \textbf{0.386  $\pm$ 0.002} \\
 distiluse-base-multilingual-cased-v2 &                              0.308 &                  \textbf{0.357  $\pm$ 0.0} &                              0.345 &                \textbf{0.389  $\pm$ 0.003} \\
paraphrase-multilingual-MiniLM-L12-v2 &                              0.261 &                 \textbf{0.332  $\pm$ 0.001} &                              0.281 &               \textbf{0.364  $\pm$ 0.004} \\
            msmarco-distilbert-cos-v5 &                              0.232 &                \textbf{0.304  $\pm$ 0.002} &                              0.262 &                \textbf{0.333  $\pm$ 0.005} \\
           multi-qa-mpnet-base-cos-v1 &                              0.274 &                 \textbf{0.324  $\pm$ 0.002 }&                              0.317 &                \textbf{0.354  $\pm$ 0.001} \\
               text-embedding-3-small &                              0.374 &                 \textbf{0.382  $\pm$ 0.002} &                              0.416 &                \textbf{0.422  $\pm$ 0.005} \\
               text-embedding-3-large &                              0.366 &                 \textbf{0.382  $\pm$ 0.007} &                              0.428 &                \textbf{0.429  $\pm$ 0.012} \\
                        voyage-3-lite &                             \textbf{0.359} &                 0.322  $\pm$ 0.005 &                              \textbf{0.400 } &                0.352  $\pm$ 0.002 \\

\bottomrule
\end{tabular}

}
\caption{\textbf{Evaluation on Task 2: Semantic similarity with graded relevance.} The table presents correlation between cosine similarity  between human \& machine summaries and relevance(ground truth) provided by human evaluators \label{tab:eval_machine_summary_downstream}.  Mean and standard error are reported based on results from two separate LLM runs for anonymization.}
\end{table*}

\section{Can anonymization help in down-stream tasks that use similarity from text-embedding models?}
\looseness=-1
In this section, we investigate the performance of the anonymized text embeddings on two downstream tasks.  Both the tasks are based on obtaining a similarity score between pieces of texts. These tasks are primarily based upon semantic similarity which find applications in areas such as information retrieval, clustering, plagiarism detection, question answering etc.~\citep{reimers2019sentence}. The two tasks that we evaluate on differ in various aspects such as the nature of the task, evaluation methodology, the judgment score available, etc. 
On both these tasks, our experiments show that embeddings based on anonymized text can significantly help in downstream tasks.

\subsection{Task 1: Semantic Similarity Between Query and Text-Pairs with Binary Labels.}
\label{sec:exp_sts_binary}
Recall from Sec.~\ref{sec:benchmarking} that altering only the names/locations in two otherwise identical stories/paragraphs significantly impacted their text embeddings. In this section, we investigate whether anonymization technique proposed in Sec.~\ref{sec:method} can effectively mitigate this type of bias. Towards this, we explore the Semantic Similarity Task (STS).

Semantic similarity seeks to determine the degree to which two pieces of text convey similar meaning~\cite{muennighoff2022mteb, reimers2019sentence}. This goes beyond simple word matching, aiming to understand the underlying meaning within the text. In today's era of deep learning~\cite{reimers2016task, muennighoff2022mteb}, achieving accurate semantic similarity relies heavily on high-quality embeddings, which represents sentences as dense vectors in a continuous space. 

\looseness=-1
In this experiment we investigate whether the text-embeddings are able to capture the semantic nuances within the text or are they biased towards names? Ideally, a good embedding model should be able to differentiate reasonably well between two stories/paragraphs which have very different themes even if they contain same names. To investigate this, we create a dataset of $10$ paragraph triplets. Each triplet includes a \textit{query} paragraph, a \textit{positive} paragraph that is \textit{highly semantically similar} but with distinct person and location names, and a \textit{negative} paragraph that is semantically \textit{dissimilar} to the query text but has same person names/location names as in query text. For each triplet, (\textit{query}, \textit{positive}) pair is assigned a label 1(positive) and \textit{(query, negative)} pair is assigned a label 0(negative). Two sample examples can be found in Table~\ref{tab:sim_change_open_ai} in the the rows marked as \textit{Original}. The entire set of generated triplets with labels are present in Appendix~\ref{sec:dataset_sts}.  We evaluate the performance of different models on the STS task using \textit{AUC ROC score} between cosine similarity scores of embeddings and the ground truth.

\paragraph{Peformance on Semantic Similarity.}
 Tab.~\ref{tab:sts_generated_with_reference_metric} presents the AUC-ROC scores for different models on the STS task. The  results indicate that the AUC scores for the majority of models are significantly below $0.5$. This finding suggests a critical issue, as even a random classifier would be expected to achieve an AUC score of approximately $0.5$. The fact that most of the AUC is much below $0.5$ suggests that the cosine similarity based ranking got the ordering wrong! 
 Gemini's AUC is better than random, however, it also gets improved significantly after anonymization. Such low AUC scores strongly imply that the embeddings used in these models are primarily capturing identity-related information, leading to a significant bias in the model's embeddings and predictions. Next, we observe that the AUC-ROC results post anonymization. We see that anonymization can improve the model's capacity to grasp the core semantic meaning in the text as reflected in the significantly higher AUC-ROC numbers(closer to 1). Additionally, it is important to note that all models attain high AUC scores when all stories share identical names.  This indicates that the models can effectively distinguish between sentences conveying the same or different meanings when identity information remains constant. The aforementioned observations highlights that anonymization is crucial to avoid situations where semantically equivalent paragraphs are assigned unique embeddings solely based on the presence of identity information (such as names). Conversely, it's essential that when texts have significant semantic variations, even if they contain identical identity information, their embeddings are able to able to capture it.

 \looseness=-1
\paragraph{Examples of similarity post-anonymization.}
In Tab.~\ref{tab:sim_change_deloc_msmarco}, we show some instances of how similarity values between embeddings change between \textit{(query, positive)} pair and  \textit{(query, negative)} pair post anonymization. Before anonymization, the models assigned higher similarity scores to negative pairs and lower similarity scores to positive pairs in a counterintuitive way. Anonymization resulted in the models predominantly attending to the semantic structure of the text, which is accurately reflected in similarity scores. We would like to highlight that these samples are a subset of examples used for AUC computation on the STS task in Tab.~\ref{tab:sts_generated_with_reference_metric}.

\subsection{Task 2: Semantic Similarity With Graded Human Relevance.}
\label{sec:exp_sts_graded_1_5_summ}  

In the previous task, a binary approach was employed to assess text pair similarity, categorizing text-pairs as either similar or dissimilar.  In the task proposed in this section, we employ a more refined approach for evaluation by utilizing a graded relevance scale from 1 to 5 between a pair of text. The graded scale enables a more nuanced and granular assessment of semantic similarity between pairs, providing a richer understanding of their relationship.
To evaluate this, we use the machine summary evaluation task from ~\citet{muennighoff2022mteb}, which involves automatically assessing the relevance of machine-generated summaries, commonly assessed by calculating the semantic similarity between the embeddings of the summary and the original document/human summaries.

For this task, we follow the same evaluation setup as~\citet{muennighoff2022mteb} which we describe next. We use the SummEval dataset~\cite{fabbri2021summeval, muennighoff2022mteb} with $100$ text samples, each containing $16$ machine and $10$ human summaries. Human relevance scores ($1{-}5$) are assigned to each machine summary. We first obtain summary embeddings using text-embedding models for each machine summary and human summary in all $100$ samples. Without loss of generality, for a given text sample out of the $100$ samples, for each machine summary $\{m_i \mid 1 \leq i \leq 16\}$, we get its predicted score based on its maximum cosine similarity to any human summary $\{h_j \mid 1 \leq j \leq 10\}$ within the same text sample i.e $machine\_pred\_score(m_i)= max_{1 \leq j \leq 10}\; cos\_sim(m_i, h_j)$. This yields $16$ machine summary quality predicted scores for each sample i.e $1$ predicted score for each machine summary. Further, as mentioned earlier, we have a human relevance score assigned to each machine summary. Overall, across all text samples, we get $1600$ \textit{machine summary predicted scores} and its corresponding \textit{human relevance scores}.  We then correlate these two scores using Pearson and Spearman coefficients~\cite{muennighoff2022mteb}. Higher correlations indicate better alignment between model-assigned scores and human judgments, suggesting more reliable evaluation.
\vspace{-0.1in}

\paragraph{Impact of Anonymization}

Table~\ref{tab:eval_machine_summary_downstream} shows that post-anonymization, the performance of various text-embedding models significantly improves in predicting graded human-rated summary quality. Spearman and Pearson correlation coefficients increase substantially, indicating that the model's assessment of summary quality after anonymization better aligns with human evaluations. This improvement is substantial, with some models like \textit{all-distilroberta-v1} showing a performance increase of around $30\%$.

\noindent
In summary, the results of both downstream tasks demonstrate a substantial enhancement in the semantic similarity post-anonymization.

\section{Conclusion}
In this work, we highlight the bias in text embeddings stemming from the presence of names in the text. We showed concrete examples, over multiple text-embedding models, that similarities between embeddings can be dominated by names in the text rather than the semantic meanings of the text. We then proposed a method to mitigate bias by performing anonymization at inference time. This involved the removal of names from the text and using the anonymized text to create the embeddings. Our findings demonstrate that anonymized text embeddings significantly outperform deanonymized text embeddings on tasks involving semantic similarity. While we proposed one way to mitigate the issue through anonymization, a deeper question that remains is: how  to train text-embedding models such that the embeddings capture the semantics more than the names in the text? 

\section{Limitations}
Below we discuss the limitations of the proposed work.
\begin{enumerate}
    \item In this work we focused on evaluating/mitigating name bias in text-embedding models using texts from English language. The work presented here does not cover other languages. Further, the work also does not cover name bias issues arising in multi language texts.

    \item While our proposed anonymization solution enhances thematic similarity, it is not ideal for situations requiring the preservation of identity that we are removing through anonymization.   A partial and straightforward solution might involve anonymizing only non-critical identifying information depending upon the use-case.  Many real world use cases may require dynamically balancing identity and thematic preservation to suit the specific needs of each use case.

    \item In our work, we adopted similarity between text-embeddings as a proxy for their semantic similarity. While commonly used, it is still an estimate of semantic similarity and may overlook deeper semantic relationships that require reasoning. A limitation of this work is that we capture thematic similarity only to the extent that it is captured by the cosine similarity (and the Euclidean distance similarity is studied in the Appendix).

\end{enumerate}

\bibliography{custom}

\clearpage
\appendix

\section{Names used for perturbation in  Benchmarking}
Table~\ref{tab:names_used_benchmark} presents the universe of names used for perturbation in the benchmarking experiment in Sec.~\ref{sec:benchmarking}.  These names represent a diverse range of geographies.
\begin{table*}[ht!]
\scalebox{0.8}{
\begin{tabular}{|c|p{17.5cm}|}
\hline
\centering
Person names&  
    Aaron, Adrian, Aiden, Akira, Alex, Alexander, Alfred, Anders, Andreas, Andrew, Anthony, Archer, Arthur, Ayden, Benjamin,    Bernard, Blake, Boris, Bradley, Brandon, Brayden, Brian, Caleb, Cameron, Carlos,    Carl, Charles, Charlie, Christopher, Connor, Cooper, Daichi, Daniel, David, Dean,    Dennis, Dylan, Edward, Elijah, Elliot, Emil, Eric, Ethan, Evan, Ezra, Fabian, Felix, Finn, Francis, Gavin, George, Giovanni, Gregory, Haakon, Han, Harry, Hayden, Henry, Hiroki, Hugo, Hunter, Ian, Isaac,    Ivan, Jack, Jacob, Jake, James, Jason, Jasper, Jayden, Jeremy,
    Jesse, Jin, Joaquim, Johan, John, Jonathan, Jordan, Joseph, Joshua, Juan,
    Kai, Kaiden, Kazuma, Keanu, Ken, Kenneth, Kevin, Liam, Logan, Lucas,
    Luis, Luke, Luka, Magnus, Mark, Martin, Mateo, Matthew, Max, Maximilian,
    Michael, Mikael, Nathan, Nathaniel, Nicolas, Noah, Oliver, Oscar,
    Owen, Pablo, Patrick, Paul, Pedro, Peter, Phillip, Phoenix, Rafael, Rajiv,
    Ralf, Ramón, Raphael, Ravi, Raymond, Reuben, Richard, Robert, Robin, Rohan,
    Roland, Ronan, Ryan, Samuel, Santiago, Sebastian, Sean, Silas, Simon, Stefan,
    Stephen, Thomas, Timothy, Tyler, Victor, Vincent, Walter, William, Xavier, Yan,
    Yang, Yao, Youssef, Zachary, Zane, Zayd, Zephyr, Zidan, Zinedine, Zubin,
    Alistair, Anders, Arjun, Arthur, Axel, Bartosz, Ben, Björn, Bruno, Caleb,
    Caoimhín, Cillian, Cormac, Daisuke, Damien, Darius, Deniz, Dorian, Eamon,
    Emile, Enzo, Fionn, Florian, Gabriel, Gideon, Gustaf, Hassan, Héctor, Igor,
    Ishaan, Ivan, Jasper, Kai, Leo, Levi, Liam, Luca, Lucian, Luis,
    Magnus, Marcel, Matteo, Max, Milan, Noah, Oliver, Oscar, Otto, Pavel,
    Quentin, Rafael, Ravi, Rémy, Ren, Robin, Samuel, Santiago, Sebastian, Silas,
    Soren, Theo, Thomas, Tristan, Viktor, William, Xavier, Yannik, Zane, Aditya, Ajeet, Ajit, Akash, Amar, Amit, Arjun, Aryan,    Ashish, Avinash, Bharat, Bhuvan, Chirag, Darshan, Dev, Dheeraj, Dhruv, Gaurav,    Harsh, Harsha, Hemant, Ishan,Shubham, Karan, Karthik, Kumar,
    Manav, Manoj, Mihir, Nikhil, Niranjan, Nivaan, Pradeep, Pranav, Raj, Rajeev,
    Rahul, Ramesh, Ranjit, Ravi, Rohan, Rohit, Roop, Sachin, Sandeep, Sanjay,
    Sanket, Sarthak, Satish, Shaan, Shahrukh, Shankar, Sharad, Shivam, Siddhant, Siddharth,
    Soham, Somesh, Suresh, Tejas, Uday, Varun, Vijay, Vikram, Vinay, Vishal,
    Yash, Yogesh, Yuvraj, Adil, Amine, Anas, Fayçal, Hakim,
    Hicham, Mazen, Mehdi, Nassim, Rafik, Sami, Sofiane, Tarik, Yacine, Yassine,
    Abiodun, Ade, Adekunle, Adewale, Ayodeji, Chidi, Chijioke, David, Ebuka, Emeka,
    Godwin, Ikechukwu, Ikenna, Kolade, Kunle, Nonso, Obinna, Olamide, Olusegun, Onyeka,
    Paul, Peter, Samuel, Taiwo, Uche, Victor, Yemi, Yinka, Aiden, Callum,
    Connor, Declan, Dylan, Eoghan, Finn, Jack, James, Jamie,
    Jason, Jayden, Kian, Liam, Logan, Lucas, Luke, Mason, Max, Michael,
    Noah, Oliver, Oscar, Rory, Ryan, Samuel, Sean, Thomas, William, Charlie,
    Freddie, George, Harry, Jacob, Leo, Oliver, Oscar, Teddy, Arthur, Freddie,
    George, Harry, Jacob, Leo, Oliver, Oscar, Teddy, Aiden, Alexander, Charlie,
    Ethan, Jacob, James, Leo, Mason, Michael, Noah, Oliver, William, Benjamin,
    Charlie, Jacob, Leo, Oliver, Oscar, Thomas, William, Aiden, Charlie, Ethan,
    Jacob, Leo, Oliver, Oscar, Thomas, William, Shrey,Venkatesh,Nguyen,Vishwanathan ,    Priya, Patricia, Jennifer, Linda, Barbara, Susan, Camille, Sophie, Julie, Claire, Yuki, Sakura, Hana, Aiko,  Emi, Li, Xiao, Mei, Fang, Jing, Maria, Ana, Isabel, Carmen,
    Dolores, Amina, Layla, Nadia, Olga, Irina, Svetlana,
    Ekaterina, Giulia, Francesca, Anna, Elena, Heidi, Greta, Lena, Marta, Sofia,
    Valentina, Martina, Paula, Clara, Laura, Mia, Emily, Sophia, Charlotte,
    Anita, Kavita, Lalita, Meena, Lucy, Megan, Hannah, Jessica, Amelia,
    Chloe, Manon, Lea, Elodie, Amandine, Haruka, Miyu, Rina, Yuna, Nao,
    Chen, Hua, Ling, Qing, Yan, Lucia, Pilar, Rosa, Nour, Sara,
    Hiba, Mona, Rania, Anastasia, Natalia, Daria, Polina, Vera, Mariana, Gabriela,
    Beatriz, Rafaela, Camila, Juliana, Evelyn, Amanda, Milla, Ines, Susana,
    Leonor, Bianca, Livia, Helena, Marina, Fernanda, Eduarda, Victoria, Andressa, Denise,
    Raquel, Isis, Elisa, Julia, Luana, Milena, Yasmin, Alessandra, Claudia, Veronica,
    Larissa, Bia, Silvia, Vanessa, Leticia, Nicole, Daniele, Eva, Alice, Milena,
    Leonie, Mila, Lisa, Sarah, Emma, Helena, Anja,
     Tina, Ingrid, Lucija,    Noor, Samira, Dana, Kalila, Arwa, Eman, Latifa, Nahla, Sang,
    Jin, Hye, Soo, Mi, Eun, Yeon, Ji, Sun, Abeba,
    Hadia, Fatou, Maimouna, Nia, Asha, Kamaria, Mira, Joan,
    Fiona, Leanne, Orla, Ava, Siobhan, Niamh, Sienna, Poppy, Lara,
    Freya, Florence, Rosie, Summer, Ivy,Sunidhi, Amara, Chidinma, Ngozi,
    Sunaina, Matilda, Harper, Willow, Aarushi, Ananya, Bhavna, Chandni, Deepa,
    Esha, Hina, Sneha, Jaya, Kiran, Lata, Maya, Nisha, Shrishti, Isabella, Saanvi, Drishti \\
\hline
Country Names & Afghanistan, Albania, Algeria, Andorra, Angola, Antigua and Barbuda, Argentina, Armenia, Australia, Austria,
    Azerbaijan, Bahamas, Bahrain, Bangladesh, Barbados, Belarus, Belgium, Belize, Benin, Bhutan,
    Bolivia, Bosnia and Herzegovina, Botswana, Brazil, Brunei, Bulgaria, Burkina Faso, Burundi, Cabo Verde, Cambodia,
    Cameroon, Canada, Central African Republic, Chad, Chile, China, Colombia, Comoros, Congo, Costa Rica,
    Croatia, Cuba, Cyprus, Czech Republic, Denmark, Djibouti, Dominica, Dominican Republic, Ecuador, Egypt,
    El Salvador, Equatorial Guinea, Eritrea, Estonia, Eswatini, Ethiopia, Fiji, Finland, France, Gabon,
    Gambia, Georgia, Germany, Ghana, Greece, Grenada, Guatemala, Guinea, Guinea-Bissau, Guyana,
    Haiti, Honduras, Hungary, Iceland, India, Indonesia, Iran, Iraq, Ireland, Israel,
    Italy, Jamaica, Japan, Jordan, Kazakhstan, Kenya, Kiribati, Kuwait, Kyrgyzstan, Laos,
    Latvia, Lebanon, Lesotho, Liberia, Libya, Liechtenstein, Lithuania, Luxembourg, Madagascar, Malawi,
    Malaysia, Maldives, Mali, Malta, Marshall Islands, Mauritania, Mauritius, Mexico, Micronesia, Moldova,
    Monaco, Mongolia, Montenegro, Morocco, Mozambique, Myanmar, Namibia, Nauru, Nepal, Netherlands,
    New Zealand, Nicaragua, Niger, Nigeria, North Korea, North Macedonia, Norway, Oman, Pakistan, Palau,
    Panama, Papua New Guinea, Paraguay, Peru, Philippines, Poland, Portugal, Qatar, Romania, Russia,
    Rwanda, Saint Kitts and Nevis, Saint Lucia, Saint Vincent and the Grenadines, Samoa, San Marino, Saudi Arabia, Senegal, Serbia  \\
\hline
\end{tabular}}
\caption{Universe of names used for replacement in benchmarking\label{tab:names_used_benchmark}.}
\end{table*}


\section{Bias measurement with only person name perturbations}
\label{sec:person_name_perturbation}
In the benchmarking study in Sec.~\ref{sec:benchmarking},  we investigated the divergence in text embeddings when person names and locations were perturbed. In this section, we examine the impact of replacing only person names on the text embeddings.

\subsection{Perturbations of only person names}
In this study, we only perturb person names and keep the location names unchanged to understand the impact of only perturbing person names. As shown in Table~\ref{tab:cmu_book_only_person_replace}, performing only person name perturbations on book plots also reveals a significant drop in cosine similarity across all evaluated models.
\label{sec:person_name_perturbation_only_country_exists}

\begin{table}[h!]
\centering
\scalebox{0.78}{\begin{tabular}{ll}
\toprule
                           Model Name &  \makecell{Cosine sim \\ per perturbation
pair} \\ 

\midrule
                all-mpnet-base-v2 &      0.815 $\pm$ 0.0001 \\
                 all-distilroberta-v1 &                 0.821 $\pm$ 0.0001 \\
                     all-MiniLM-L6-v2 &                 0.749 $\pm$ 0.0002 \\
                               gemini &                     0.91 $\pm$ 0.0 \\
           multi-qa-distilbert-cos-v1 &                 0.787 $\pm$ 0.0001 \\
              paraphrase-MiniLM-L6-v2 &                 0.773 $\pm$ 0.0003 \\
 distiluse-base-multilingual-cased-v1 &                 0.843 $\pm$ 0.0002 \\
 distiluse-base-multilingual-cased-v2 &                 0.848 $\pm$ 0.0002 \\
paraphrase-multilingual-MiniLM-L12-v2 &                  0.79 $\pm$ 0.0003 \\
            msmarco-distilbert-cos-v5 &                 0.752 $\pm$ 0.0001 \\
           multi-qa-mpnet-base-cos-v1 &                 0.795 $\pm$ 0.0001 \\
                        voyage-3-lite &                 0.821 $\pm$ 0.0001 \\
                               
\bottomrule
\end{tabular}}
\caption{\textbf{Bias Measuremenent on CMU Books dataset with perturbation of person names only.} For each show, we create $K{=}20$ perturbations by
replacing person names. We compute the average cosine similarity for each perturbation pair and the standard error. The country/city/town names remain unchanged. \label{tab:cmu_book_only_person_replace}.}
\end{table}

\subsection{Person name perturbations on text samples without mention of country/city/town names}
\label{sec:person_name_no_country_samples}
In this section, we investigate impact of person name perturbations when using samples which don't have mention of any country/city/town etc. names. The objective is to minimize the impact of these variables and study divergence solely w.r.t person names. As shown in Table~\ref{tab:cmu_book_no_location_only_person_replace}, benchmarking on the CMU Book dataset on samples without having any mention of country/city/town etc. reveals a significant drop in cosine similarity across all evaluated models when only person names are perturbed.

\begin{table}[h!]
\centering
\scalebox{0.78}{
\begin{tabular}{ll}
\toprule
                          Model Name &  \makecell{Cosine sim \\ per perturbation
pair}    \\
\midrule
                    all-mpnet-base-v2 &                 0.796 $\pm$ 0.0002 \\
                 all-distilroberta-v1 &                 0.803 $\pm$ 0.0002 \\
                     all-MiniLM-L6-v2 &                 0.731 $\pm$ 0.0003 \\
                               gemini &                 0.906 $\pm$ 0.0001 \\
           multi-qa-distilbert-cos-v1 &                 0.766 $\pm$ 0.0002 \\
              paraphrase-MiniLM-L6-v2 &                 0.758 $\pm$ 0.0004 \\
 distiluse-base-multilingual-cased-v1 &                 0.825 $\pm$ 0.0003 \\
 distiluse-base-multilingual-cased-v2 &                 0.828 $\pm$ 0.0003 \\
paraphrase-multilingual-MiniLM-L12-v2 &                  0.77 $\pm$ 0.0004 \\
            msmarco-distilbert-cos-v5 &                 0.747 $\pm$ 0.0002 \\
           multi-qa-mpnet-base-cos-v1 &                 0.778 $\pm$ 0.0002 \\
                        voyage-3-lite &                  0.81 $\pm$ 0.0001 \\
\bottomrule
\end{tabular}}
\caption{\textbf{Bias Measuremenent on CMU Books dataset on samples without mention of country/city/town names.} Perturbation  of person names only. For each show, we create $K{=}20$ perturbations by
replacing person names. We compute the average cosine similarity for each perturbation pair and the standard error.\label{tab:cmu_book_no_location_only_person_replace}}
\end{table}

\begin{table*}[h!]
\scalebox{0.8}{
\begin{tabular}{|c|p{17cm}|}
\hline
Country & Person Names \\ \hline
 France &  
   Max, Tom, Léo, Noé, Paul, 
    Jules, Hugo, Arthur, Louis, Clément,
    Jean-Baptiste, Jean-Pierre, Jean-Paul, Charles-Henri, François-Xavier, 
    Constantin, Gaspard, Côme, Yanis, Kilian,
    Maël, Thibault, Raphaël, Jérémie, Vincent,
    Antoine, Pierre, Louis, Jacques, Baptiste,
    Émile, Gustave, Henri, Laurent, Marcel,
    Nicolas, Olivier, Pascal, Quentin, Rémi,
    Sébastien, Théodore, Ulysse, Valentin, Wilfried,
    Xavier, Yves, Zacharie, Adrien, Bernard, Eva,  Zoé,  Jade,  Lou,
    Alice, Chloé, Léa,  Lina, Louise, 
    Éléonore, Solène, Héloïse, Camille, 
    Marie,  Jeanne,  Sophie,  Claire,  Isabelle, 
    Ambre,  Lilou,  Maëlys, Victoire,  Clémence,
    Valentine, Juliette, Aurélie, Angélique, Amandine,
    Brigitte, Catherine, Delphine, Édith, Fanny,
    Gabrielle, Hélène, Inès, Joséphine, Karine,
    Laure, Manon, Nathalie, Océane, Pascale,
    Quitterie, Rosalie, Stéphanie, Thérèse, Ursule\\
\hline
    India & Aarav, Aditya, Aryan, Ayush, Dev, 
    Ishaan, Ramesh, Krishna, Mihir, Rohan, 
    Sahir, Samarth, Shaurya, Vihaan, Vrijesh, 
    Aakash, Advait, Vinayak, Atharv, Venkatesh, 
    Dhruv, Eshan, Hrithik, Kabir, Karan, 
    Krish, Mahesh, Nakul, Pranav, Rudra, 
    Siddharth, Soham, Tanmay, Uday, Vaibhav, 
    Vedant, Vikram, Yash, Yuvraj,Sachin, 
    Ahaan, Gaurav, Arjun, Daksh, Devansh, 
    Ishan, Vishwanathan, Mayank, Parichay, Krishnanshu, 
    Sahir, Rishi, Samyak, Brajesh, Vivaan, 
    Ayan, Rudra,Rakesh, Zain,  Aarohi, Bhavya, Charvi, Devika, Eshani,
    Falguni, Garima, Harini, Ishita, Jahnvi,
    Kavya, Lavanya, Madhavi, Niharika, Ojasvi,
    Prisha, Qara, Radhika, Saanvi, Tara,
    Urvashi, Vanya, Wamika, Xara, Yamini,
    Zara, Anvi, Bhumika, Chaitali, Dharini,
    Ekta, Fiza, Gauri, Himani, Ira,
    Jiya, Kriti, Lata, Meera, Nisha,
    Oviya, Pallavi, Rhea, Sakshi, Tanisha,
    Uma, Vaidehi, Yashika, Zaina, Aditi \\ \hline

    Spain &

    Mateo, Santiago, Lucas, Marcos, Daniel,
    David, Samuel, Benjamín, Ezequiel, Noé,
    Salvador, Ismael, Aarón, Elías, Jonás,
    Jeremías, Iker, Unax, Aitor, Ander,  
    Martín, Rodrigo, Fernando, Alfonso, Enrique,
    Felipe, Carlos, Javier, Jorge, Luis,
    Antonio, José, Juan, Manuel, Pedro,
    Francisco, Ignacio, Rafael, Víctor, Álvaro,
    Diego, Gabriel, Miguel, Pablo, Ricardo,
    Sergio, Tomás, César, Gonzalo, Leonardo,
    Emiliano, Matías, Nicolás, Sebastián, Thiago,   Sofía, Camila, Valentina, Martina, Emilia,
    Emma, Olivia, Luna, Zoe, Mia,
    Isabella, Victoria, Sara, Lucía, María,
    Laura, Paula, Andrea, Ana, Elena,
    Carmen, Alba, Carla, Daniela, Julia,
    Natalia, Ximena, Aitana, Noa, Mía,
    Isabel, Beatriz, Blanca, Clara, Inés,
    Irene, Marta, Patricia, Rocío, Silvia,
    Teresa, Verónica, Alicia, Amelia, Ángela,
    Aurora, Bárbara, Carolina, Dolores, Eva,
    Gloria, Lidia, Lorena, Mónica, Nuria,
    Olga, Raquel, Sandra,Xiomara, Yamile \\

\hline
\end{tabular}}
\caption{Universe of names for country wise name replacement in benchmarking experiments in Sec.~\ref{sec:person_name_perturbation}\label{tab:names_used_benchmark_country_wise}}
\end{table*}

\begin{table}[h!]
\centering
\scalebox{0.78}{
\begin{tabular}{ll}
\toprule
                           Model Name &    \makecell{Cosine sim \\ per perturbation
pair}   \\
\midrule
                    all-mpnet-base-v2 &                 0.842 $\pm$ 0.0002 \\
                 all-distilroberta-v1 &                 0.852 $\pm$ 0.0002 \\
                     all-MiniLM-L6-v2 &                 0.784 $\pm$ 0.0002 \\
                               gemini &                     0.93 $\pm$ 0.0 \\
           multi-qa-distilbert-cos-v1 &                  0.82 $\pm$ 0.0002 \\
              paraphrase-MiniLM-L6-v2 &                 0.806 $\pm$ 0.0004 \\
 distiluse-base-multilingual-cased-v1 &                 0.837 $\pm$ 0.0003 \\
 distiluse-base-multilingual-cased-v2 &                 0.838 $\pm$ 0.0003 \\
paraphrase-multilingual-MiniLM-L12-v2 &                  0.82 $\pm$ 0.0003 \\
            msmarco-distilbert-cos-v5 &                 0.799 $\pm$ 0.0002 \\
           multi-qa-mpnet-base-cos-v1 &                 0.815 $\pm$ 0.0002 \\
                        voyage-3-lite &                 0.847 $\pm$ 0.0001 \\
                        
\bottomrule
\end{tabular}}
\caption{\textbf{Bias Measurement: Names from same country.} Perturbation of person names and replacing them with names from \textbf{\textit{Spain}}. We used CMU Book dataset for this experiment and set number of perturbations $K{=}20$. We used samples without mention of country/city/town/other location names, nationality etc.\label{tab:same_culture_perturb_Spain} } 
\end{table}

\begin{table}[h!]
\centering
\scalebox{0.78}{
\begin{tabular}{ll}
\toprule
                           Model Name &   \makecell{Cosine sim \\ per perturbation
pair} \\
\midrule
                    all-mpnet-base-v2 &                  0.84 $\pm$ 0.0002 \\
                 all-distilroberta-v1 &                 0.838 $\pm$ 0.0002 \\
                     all-MiniLM-L6-v2 &                 0.757 $\pm$ 0.0003 \\
                               gemini &                    0.931 $\pm$ 0.0 \\
           multi-qa-distilbert-cos-v1 &                 0.806 $\pm$ 0.0002 \\
              paraphrase-MiniLM-L6-v2 &                  0.79 $\pm$ 0.0004 \\
 distiluse-base-multilingual-cased-v1 &                  0.83 $\pm$ 0.0003 \\
 distiluse-base-multilingual-cased-v2 &                 0.833 $\pm$ 0.0003 \\
paraphrase-multilingual-MiniLM-L12-v2 &                 0.815 $\pm$ 0.0004 \\
            msmarco-distilbert-cos-v5 &                 0.786 $\pm$ 0.0002 \\
           multi-qa-mpnet-base-cos-v1 &                  0.81 $\pm$ 0.0002 \\
                        voyage-3-lite &                 0.843 $\pm$ 0.0001 \\
\bottomrule
\end{tabular}}
\caption{\textbf{Bias Measurement: Names from same country.} Perturbation of person names and replacing them with names from \textit{\textbf{France}}. We used CMU Book dataset for this experiment and set number of perturbations $K{=}20$. We used samples without mention of country/city/town/other location names, nationality etc. \label{tab:same_culture_perturb_France}}
\end{table}

\begin{table}[h!]
\centering
\scalebox{0.78}{
\begin{tabular}{ll}
\toprule
                           Model Name &    \makecell{Cosine sim \\ per perturbation
pair}   \\
\midrule
                    all-mpnet-base-v2 &                 0.816 $\pm$ 0.0002 \\
                 all-distilroberta-v1 &                 0.828 $\pm$ 0.0002 \\
                     all-MiniLM-L6-v2 &                  0.75 $\pm$ 0.0003 \\
                               gemini &                    0.931 $\pm$ 0.0 \\
           multi-qa-distilbert-cos-v1 &                  0.79 $\pm$ 0.0002 \\
              paraphrase-MiniLM-L6-v2 &                 0.778 $\pm$ 0.0004 \\
 distiluse-base-multilingual-cased-v1 &                  0.88 $\pm$ 0.0002 \\
 distiluse-base-multilingual-cased-v2 &                 0.887 $\pm$ 0.0002 \\
paraphrase-multilingual-MiniLM-L12-v2 &                 0.796 $\pm$ 0.0004 \\
            msmarco-distilbert-cos-v5 &                  0.78 $\pm$ 0.0002 \\
           multi-qa-mpnet-base-cos-v1 &                 0.805 $\pm$ 0.0002 \\
                        voyage-3-lite &                  0.85 $\pm$ 0.0001 \\
\bottomrule
\end{tabular}}
\caption{\textbf{Bias Measurement: Names from same country.} Perturbation of person names and replacing them with names from \textbf{\textit{India}}. We used CMU Book dataset for this experiment and set number of perturbations $K{=}20$. We used samples without mention of country/city/town/other location names, nationality etc.\label{tab:same_culture_perturb_India} }
\end{table}

\subsection{Bias measurement with person name perturbations from the same geographical area}
\label{sec:person_name_same_geography}
In previous studies, we perturbed names by replacing them from a diverse set of person names. In this study we investigate whether the issue of divergence in embeddings persists when all the perturbed names are from the same geography. This study aims to minimize the impact of cultural differences in analysis in text-embeddings. Table~\ref{tab:names_used_benchmark_country_wise} shows the country wise names used for benchmarking. In tables~\ref{tab:same_culture_perturb_Spain}, ~\ref{tab:same_culture_perturb_France}, and~\ref{tab:same_culture_perturb_India}, we observe that the divergence issue persists even when the replaced names belong to the same geography. This demonstrates that the issue is not present in names from certain cultures, cross-culture, but is universal in the sense that the name bias issue occurs in a very broad sense.

\newpage
\section{Similarity Heatmaps}
\label{similarity_heatmap}
In this section, we show examples of cosine similarity heatmaps based upon embeddings generated by different text-embedding models. We use the following example:

{\color{blue} \textbf{CHARACTER\_NAME}, a seasoned physician, meticulously analyzed a patient’s intricate heart condition. He later realised she was his school friend.}

\begin{figure*}[h!]
  \centering
  \includegraphics[width=1.2\textwidth]{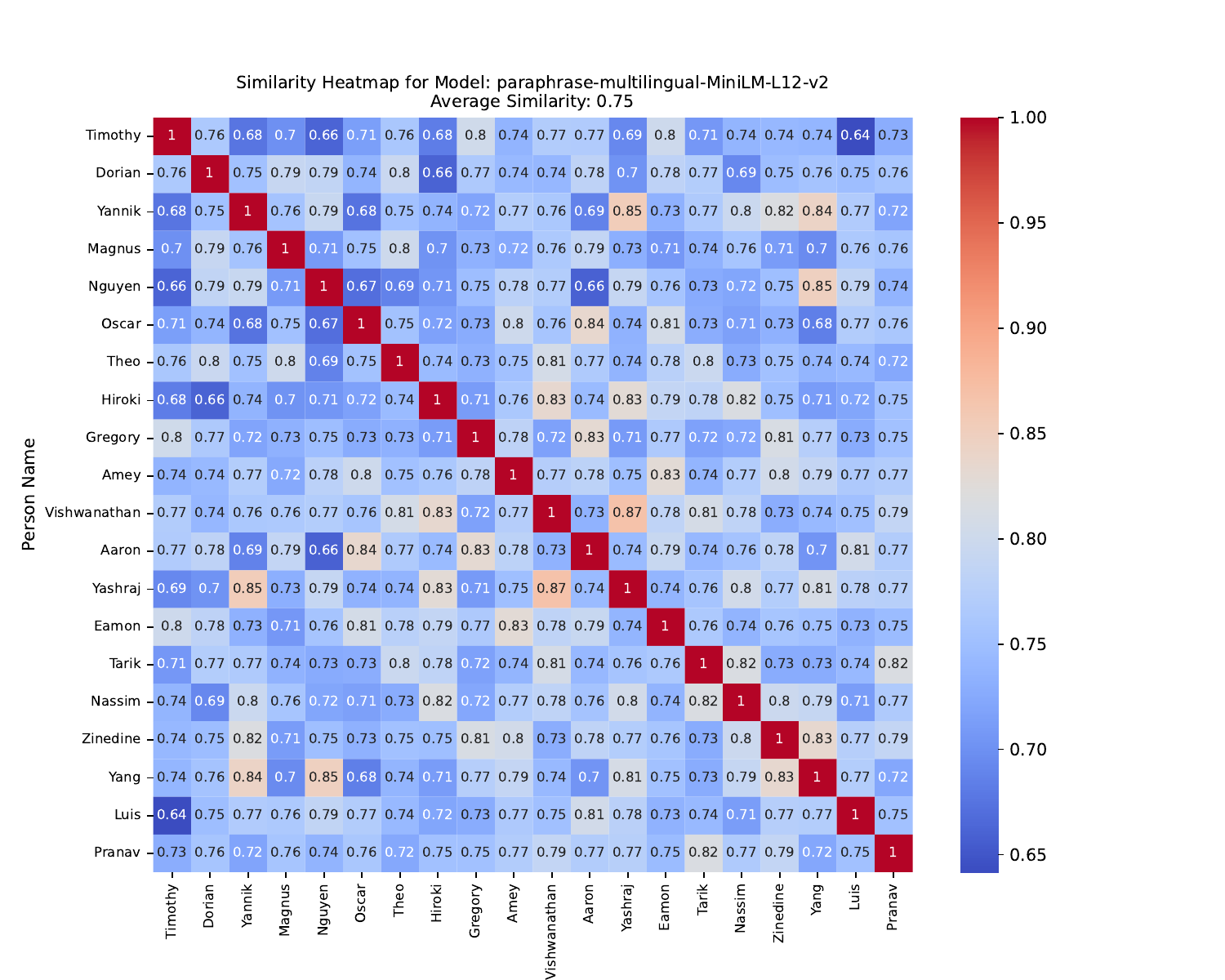}
  \caption{Cosine Similarity Heatmap with \textit{paraphrase-multilingual-MiniLM-L12} model for example in Sec.~\ref{similarity_heatmap}}
  \label{fig:heatmap_paraphrase}
\end{figure*}

\begin{figure*}[h!]
  \centering
  \includegraphics[width=1.2\textwidth]{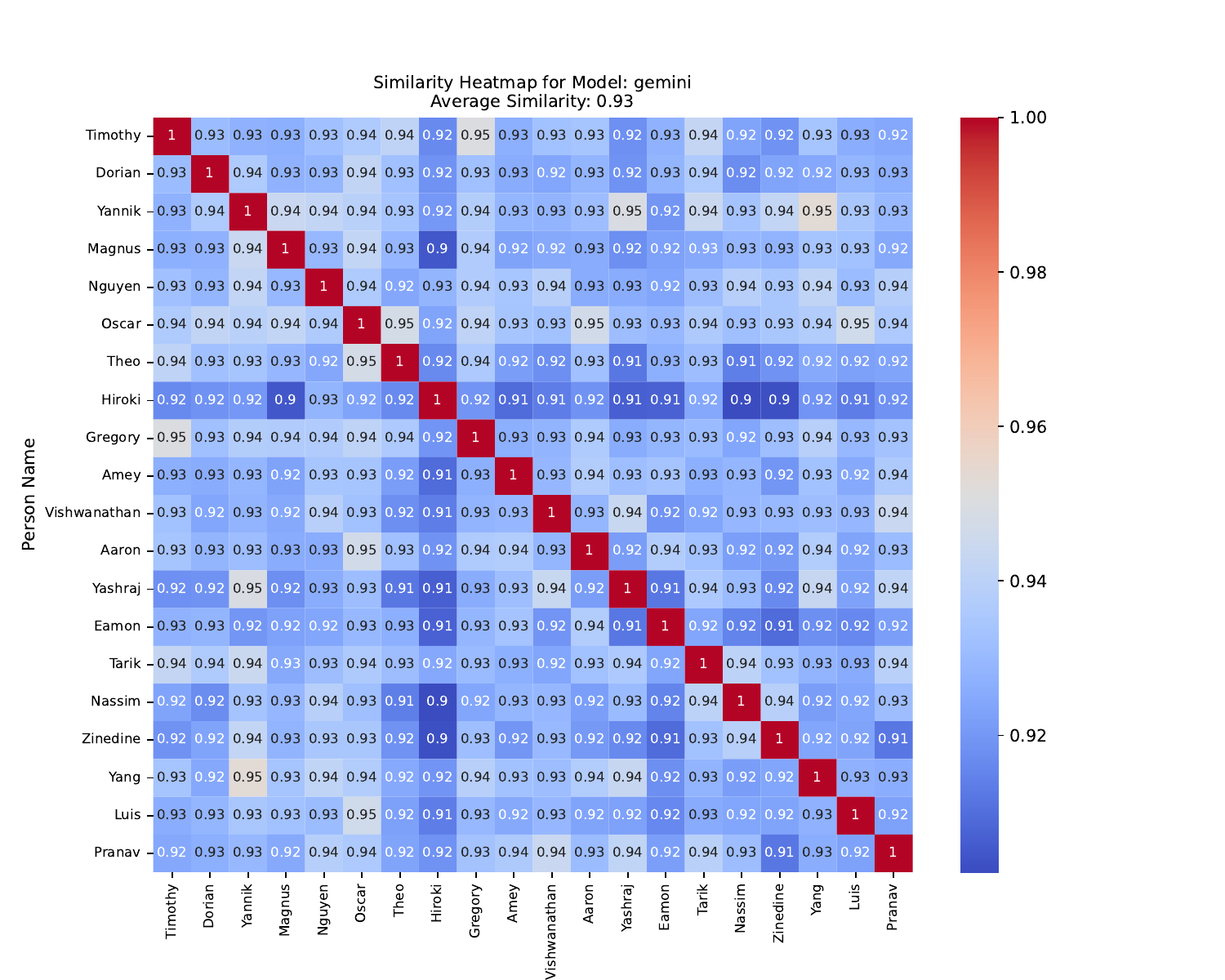}
  \caption{Cosine Similarity Heatmap with Gemini model for example in Sec.~\ref{similarity_heatmap}}
  \label{fig:heatmap_gemini}
\end{figure*}

\begin{figure*}[h!]
  \centering
  \includegraphics[width=1.2\textwidth]{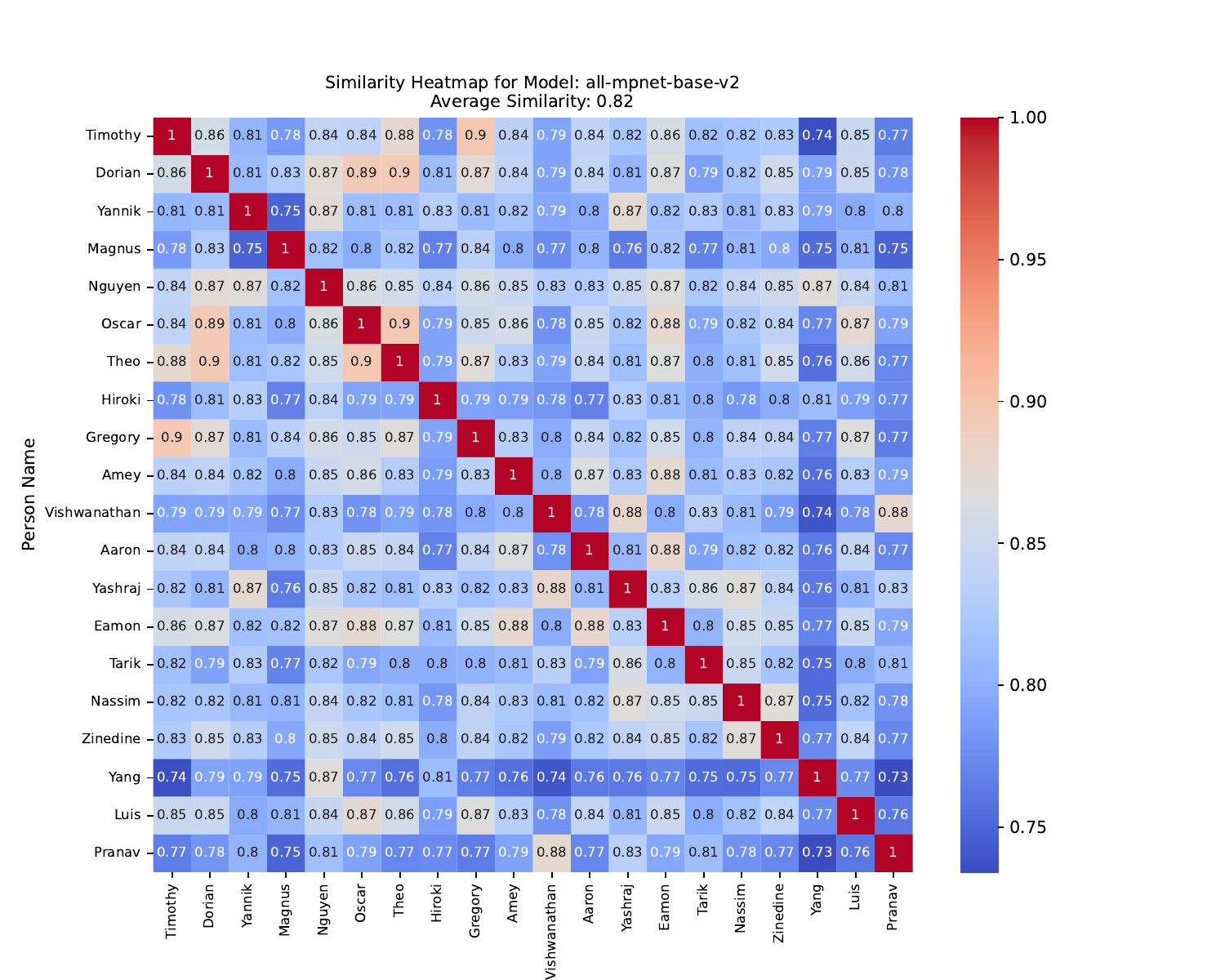}
  \caption{Cosine Similarity Heatmap with \textit{all-mpnet-base-v2} model for example in Sec.~\ref{similarity_heatmap}  \label{fig:all_heatmap_mpnet}}
\end{figure*}

\begin{figure*}[h!]
  \centering
  \includegraphics[width=1.2\textwidth]{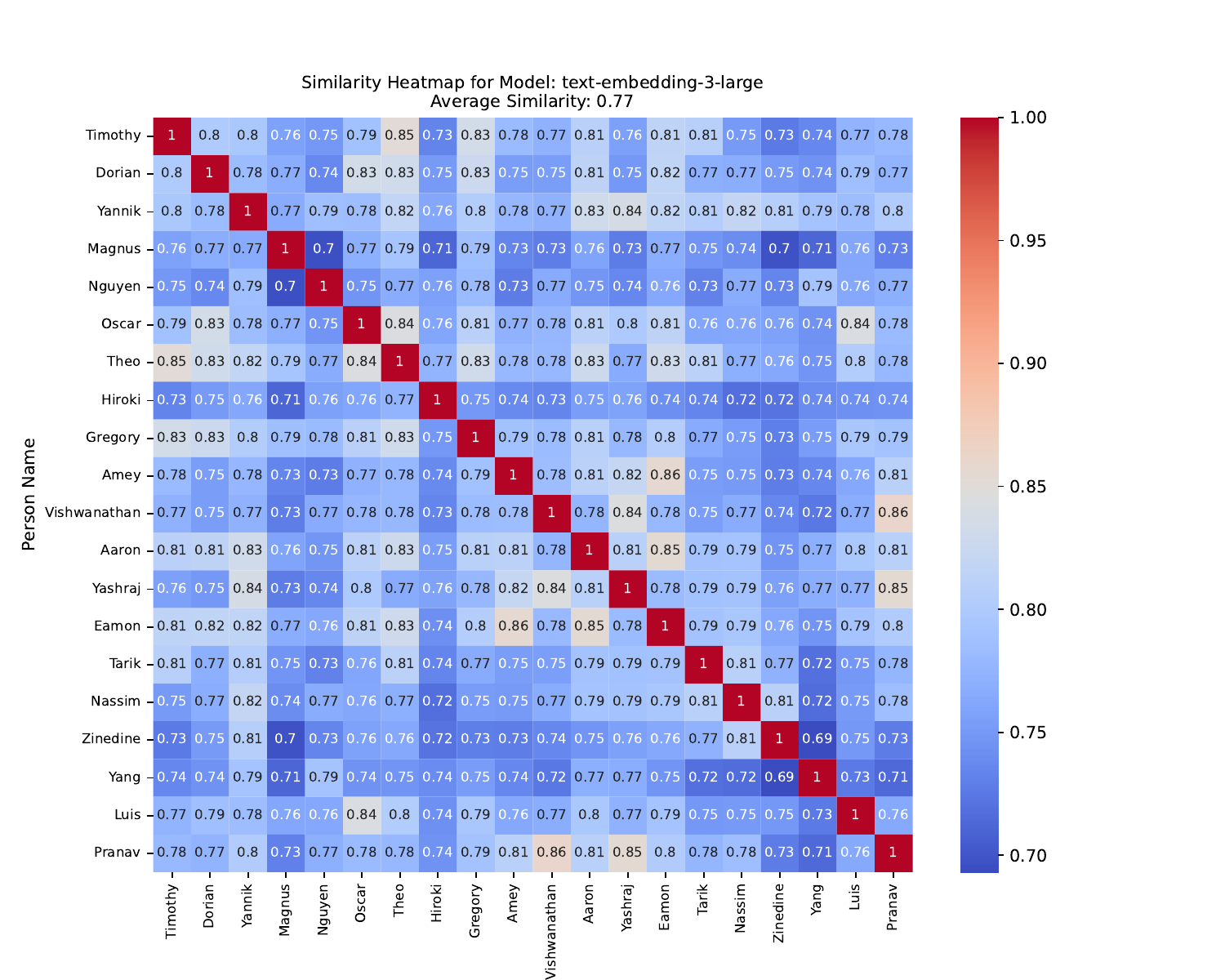}
  \caption{Cosine Similarity Heatmap with \textit{text-embedding-3-large}(Open AI) model for example in Sec.~\ref{similarity_heatmap}   \label{fig:heatmap_openai}}
\end{figure*}

To obtain different perturbations, we replace  \textbf{{\color{blue} "CHARACTER\_NAME"}} with different person names and generate embedding for each of the perturbation. The similarity heatmaps are present in  ~\cref{fig:heatmap_paraphrase,fig:heatmap_gemini,fig:heatmap_openai,fig:all_heatmap_mpnet}. The heatmaps clearly reveal that only changing the person names can significantly impact the text embeddings. This suggests that the text embedding model is highly sensitive to the specific names used within the text, even when the overall context and meaning remains completely unchanged. These kind of variations can lead to misleading results in various downstream tasks. For example, if the goal is to cluster documents into topics, changing the person names could lead to different clusters being formed, even if the underlying topics are the same. Similarly, if the text embedding model is used to classify documents as positive or negative, changing the person names could lead to different classifications being assigned, even if the overall sentiment and theme of the text remains the same. 

   \clearpage

\section{Semantic Similarity Task Dataset}
Below we present the STS dataset consisting $10$ samples used in Sec.~\ref{sec:exp_sts_binary}. Each sample is  a triplet of the form: \\ $< Query, Positive\ sample, Negative\ sample >$.
\label{sec:dataset_sts}



\begin{enumerate}
    \item \textbf{Query:} {Nikolai} and {Deborah} met on a rainy Tuesday in {New York}. The city’s hustle and bustle couldn’t dim the spark between them. {Deborah}, with her radiant smile and infectious laughter, had captured {Nikolai}'s heart from the moment he saw her. {Nikolai}, a charming and witty gentleman, returned her affection with equal fervor.
        \begin{itemize}

            \item \textbf{Positive:} {Kashvi} and {Oluwafemi} met on a rainy Tuesday in {Northampton}. The city’s bustling streets couldn’t dim the spark between them. {Kashvi}, with her radiant smile and infectious laughter, had captured {Oluwafemi}’s heart from the moment he saw her. {Oluwafemi}, a charming and witty gentleman, returned her affection with equal fervor.

            \item \textbf{Negative:} {Nikolai} and {Deborah} staying in {New Jersey}, once inseparable, were now worlds apart. {Deborah}, the trusted confidante, had betrayed {Nikolai}'s trust, revealing his secrets to their rivals. The city's hustle and bustle mirrored the chaos within {Nikolai}'s heart, as he grappled with the bitter reality of love turned treachery.
        \end{itemize}

    \item \textbf{Query:}  Alejandro quickly ran to the store to buy a cold drink. He was eager to have a glass of cold drink.
        \begin{itemize}
            \item \textbf{Positive:} Quickly, Hiroki dashed to the local market to procure some cold drinks. He was yearning for a chilled glass of cold drink.
            \item \textbf{Negative:} Alejandro has stopped buying cold drinks from market. He only drinks cold drinks made at home.
        \end{itemize}

    \item \textbf{Query:}  Mayatoshi and Alex had a deep, passionate love for each other. Their bond was unbreakable, a love that transcended all obstacles. They shared dreams, hopes, and aspirations, and their love was the foundation of their happiness.
        \begin{itemize}
            \item \textbf{Positive:} Priyanka and Yuan were deeply in love. Their affection for each other was profound and unwavering. They shared a strong connection, a love that was the source of their joy and contentment.
            \item \textbf{Negative:} Despite their intense hatred for each other, Mayatoshi and Alex were bound by a strange, twisted connection. Their animosity fueled a toxic relationship, a constant battle of wills. Their lives were intertwined, a dark, destructive dance of love and hate.
        \end{itemize}

    \item \textbf{Query:}  Amazon and Apple are two American corporations. Amazon's main business is online shopping and Apple is a phone maker giant
        \begin{itemize}
            \item \textbf{Positive:} Alibaba and Xiaomi are two Chinese corporations. Alibaba's main business is online shopping and Xiaomi is a producer of phones
            \item \textbf{Negative:} Amazon is a river in South America. Apples are not grown in the Amazon basin.
        \end{itemize}

    \item \textbf{Query:}  Ganga and Yamuna are two mighty rivers. They are lifelines for millions of people in the region.
        \begin{itemize}
            \item \textbf{Positive:} Yangtze is a mighty river. It is a long river and is the lifeline for millions of people in the region.
            \item \textbf{Negative:} Ganga and Yamuna are two sisters. They had their schooling in the region and schooling provided a lifeline for them.
        \end{itemize}

    \item \textbf{Query:}  Alice and Bob often helped each other financially. Recently, Alice lent Bob a significant sum of money. Bob promised to return it soon.
        \begin{itemize}
            \item \textbf{Positive:} Yuri and Haruto frequently helped each other out, including with money. Lately, Yuri had loaned Haruto a substantial amount of money, which Haruto assured her he’d repay promptly.
            \item \textbf{Negative:} Alice and Bob had a disagreement about money. Alice believed Bob owed her money, but Bob denied it.
        \end{itemize}

    \item \textbf{Query:}  John, a renowned lawyer, is defending his client, Mike, who is accused of a serious crime. John is determined to prove Mike's innocence and secure his acquittal.
        \begin{itemize}
            \item \textbf{Positive:} Armaan, a man falsely accused of a heinous crime, is relying on his skilled lawyer, Udit, to exonerate him. Udit is committed to presenting a strong defense and clearing Armaan's name.
            \item \textbf{Negative:} John, a cunning lawyer, is manipulating the legal system to frame Mike for a crime he did not commit. John's goal is to secure a conviction and advance his own career, regardless of the truth.
        \end{itemize}

    \item \textbf{Query:}  Dr. Alexander, a seasoned physician, meticulously analyzed patient Sarah’s intricate heart condition. He prescribed a tailored regimen of  medications and rigorous lifestyle modifications to significantly improve her cardiac health.
        \begin{itemize}
            \item \textbf{Positive:} The esteemed doctor, Dr. Yerusha, conducted a thorough assessment of patient Reyan’s complex symptoms of heart. She formulated a precise treatment plan, incorporating  medications and day to day lifestyle changes, to alleviate Reyan's debilitating heart condition.
            \item \textbf{Negative:} Dr. Alexander, a renowned doctor and surgeon, executed a high-risk heart surgical procedure on patient Sarah. After the complex operation Sarah did not recover.
        \end{itemize}

    \item \textbf{Query:}  Mr. Smith, a dedicated teacher, guided his students, including the bright young minds of Miller and Pristina, towards academic excellence.
        \begin{itemize}
            \item \textbf{Positive:} Mr. Yang, a committed educator, mentored his students, including the talented Shruti and Ren, to achieve academic success.
            \item \textbf{Negative:} Mr. Smith , a rigid and punitive teacher, often unfairly targeted mischievous students like Miller and Pristina.
        \end{itemize}

    \item \textbf{Query:}  Martinez gently examined the injured bird. He gave it food.
        \begin{itemize}
            \item \textbf{Positive:} Yohan tenderly inspected the wounded bird and gave it a meal to eat.
            \item \textbf{Negative:} The skilled hunter Martinez tracked the injured bird. He captured it for food.
        \end{itemize}

\end{enumerate}

\section{Example of Semantic Similarity post-anonymization}
In Table~\ref{tab:sim_change_open_ai}, we show impact of anonymization on STS tasks on embeddings crated by Open AI’s $text-embedding-3-small$ model. We observe that in all cases performance after anonymization is superior. Specifically, post anonymization,  we obtain relatively higher score for positive samples and lower for negative samples.
 \begin{table*}[h!]
    \centering
    \scalebox{0.9}{
    \begin{tabular}{|c|p{5cm}|p{7cm}|p{1cm}|p{1cm}|}
     & Query & Pos/Neg & Sim score & Label \\
    \hline
    \multirow{2}{*}{Original} & \multirow{2}{5cm}{Alejandro quickly ran to the store to buy a cold drink. He was eager to have a glass of cold drink.} & \textcolor{blue}{\textbf{POS:} Quickly, Hiroki dashed to the local market to procure some cold drinks. He was yearning for a chilled glass of cold drink.} & 0.66 & 1 \\
   
    & & \textcolor{red}{\textbf{NEG:} Alejandro has stopped buying cold drinks from market. He only drinks cold drinks made at home.} & 0.72 & 0 \\  \cline{3-5}
    \multirow{2}{*}{Anonymized} & \multirow{2}{5cm}{\textcolor{black}{quickly ran to the store to buy a cold drink. He was eager to have a glass of cold drink.}} & \textcolor{blue}{\textbf{POS:} Quickly,  dashed to the local market to procure some cold drinks. He was yearning for a chilled glass of cold drink.} & 0.83 & 1 \\
    & & \textcolor{red}{\textbf{NEG:} has stopped buying cold drinks from market. He only drinks cold drinks made at home.} & 0.57 & 0 \\
    \end{tabular}}

    \centering
    \scalebox{0.9}{
    \begin{tabular}{|c|p{5cm}|p{7cm}|p{1cm}|p{1cm}|}
    \hline
    \multirow{2}{*}{Original} & \multirow{2}{5cm}{Ganga and Yamuna are two mighty rivers. They are lifelines for millions of people in the region.} & \textcolor{blue}{\textbf{POS:} Yangtze is a mighty river. It is a long river and is the lifeline for millions of people in the region.} & 0.63 & 1 \\
    
    & & \textcolor{red}{\textbf{NEG:} Ganga and Yamuna are two sisters. They had their schooling in the region and schooling provided a lifeline for them.} & 0.73 & 0 \\ \cline{3-5}
    \multirow{2}{*}{Anonymized} & \multirow{2}{5cm}{and are two mighty rivers. They are lifelines for millions of people in the region.} & \textcolor{blue}{\textbf{POS:} is a mighty river. It is a long river and is the lifeline for millions of people in the region.} & 0.76 & 1 \\
    & & \textcolor{red}{\textbf{NEG:} and are two sisters. They had their schooling in the region and schooling provided a lifeline for them.} & 0.46 & 0 \\
    \hline
    \end{tabular}}
    \caption{Example demonstrating impact of anonymization on semantic similarity using embeddings created by Open AI's \textit{text-embedding-3-small} model. The text in color \textcolor{blue}{blue} and \textcolor{red}{red} refer to the positive and negative paragraphs respectively.}
    \label{tab:sim_change_open_ai}
    \end{table*}

\section{Impact of Anonymization Strategy: Removal versus Replacement}
\label{sec:anon_vs_replacement}

\begin{table*}[h!]
\centering
\begin{tabular}{|p{5cm}|p{10cm}}
\toprule
   Replace person names, organizations and locations  &  Given below text, please convert all Person names(which are Proper Nouns) to a UNIQUE ID such as CHAR\_A, CHAR\_B, CHAR\_C etc..  Keep it unique and for each UNIQUE Person name(which is a Proper Noun) use a UNIQUE ID. DO NOT KEEP THE ORIGINAL Person Names(which are Proper Nouns) in the generated paragraph text. Next, Replace all occurences  City/Country/Village/Town/River/Continent etc. names which are PROPER NOUNS to a UNIQUE ID such as LOC\_A, LOC\_B, LOC\_C, LOC\_D etc.. Next, Replace all occurences of company/organization names which are PROPER NOUNS to a UNIQUE ID such as ORG\_A, ORG\_B, ORG\_C, ORG\_D etc..   Do not replace monument/landmark names like Eiffel tower etc. Output contains the modified text only....  The text is provided below :::: \\
   \hline
\end{tabular}
\caption{Prompt for Anonymization using replacement strategy described in Sec.~\ref{sec:anon_vs_replacement} \label{tab:replacement_prompt}}
\end{table*}

This section investigates the effectiveness of remove of names vs.  replacement of names in text for anonymization. In the replacement strategy, we replace names with non-identifying placeholder names instead of removing them from text. Example: person names with 'CHAR\_{ID}', location names with 'LOC\_{ID}' etc. Here ID can be replaced with $\{A, B, C \cdots \}$ or $\{1, 2, 3 \cdots \}$ etc. The detailed prompt is present in Table ~\ref{tab:replacement_prompt}. In Table~\ref{tab:anon_vs_replacement} we demonstrate that removal of names marginally outperforms replacement in the STS task. In the context of replacement strategy, one should note that the quality of embeddings derived is sensitive to the specific replacement placeholder terms used. For instance, substituting character names with with different placeholders such as  ``CHAR\_A'' / ``CHARACTER\_B'' / ``CHARACTER\_1'' or location names with ``LOC\_1'' / ``LOC\_B'' can impact the resulting embeddings differently. In order to mitigate this sensitivity and ensure consistent results and also based upon our findings we recommend using the name removal strategy for anonymization to mitigate name bias. 

\begin{table*}[t!]
\centering
\scalebox{0.8}{\begin{tabular}{llll}
\toprule
                                Model &  \makecell{AUC ROC \\ Original} & \makecell{AUC ROC \\ Anonymization(Default)} & \makecell{AUC ROC \\ Anonymization(Replacement)} \\
\midrule

                    all-mpnet-base-v2 &                   0.19 &       0.98 $\pm$ 0.0071 &                   \textbf{1.0 $\pm$ 0.0} \\
                 all-distilroberta-v1 &                   0.36 &      \textbf{0.975 $\pm$ 0.0106} &              0.945 $\pm$ 0.0106 \\
                     all-MiniLM-L6-v2 &                   0.09 &       \textbf{0.99 $\pm$ 0.0071} &               0.97 $\pm$ 0.0071 \\
                               gemini &                   0.71 &           \textbf{1.0 $\pm$ 0.0} &                   \textbf{1.0 $\pm$ 0.0} \\
           multi-qa-distilbert-cos-v1 &                   0.07 &       \textbf{0.97 $\pm$ 0.0071} &                  0.95 $\pm$ 0.0 \\
              paraphrase-MiniLM-L6-v2 &                   0.14 &          0.98 $\pm$ 0.0 &               \textbf{0.99 $\pm$ 0.0071} \\
 distiluse-base-multilingual-cased-v1 &                   0.27 &          \textbf{0.94 $\pm$ 0.0} &              0.935 $\pm$ 0.0106 \\
 distiluse-base-multilingual-cased-v2 &                   0.26 &          \textbf{0.96 $\pm$ 0.0} &               0.94 $\pm$ 0.0212 \\
paraphrase-multilingual-MiniLM-L12-v2 &                   0.21 &          0.99 $\pm$ 0.0 &                   \textbf{1.0 $\pm$ 0.0} \\
            msmarco-distilbert-cos-v5 &                   0.10 &      \textbf{0.955 $\pm$ 0.0035} &              0.875 $\pm$ 0.0035 \\
           multi-qa-mpnet-base-cos-v1 &                   0.08 &           \textbf{1.0 $\pm$ 0.0} &              0.985 $\pm$ 0.0035 \\
               text-embedding-3-small &                   0.12 &           1.0 $\pm$ 0.0 &               0.97 $\pm$ 0.0071 \\
               text-embedding-3-large &                   0.21 &           \textbf{1.0 $\pm$ 0.0} &                   \textbf{1.0 $\pm$ 0.0} \\
                        voyage-3-lite &                   0.18 &           \textbf{1.0 $\pm$ 0.0} &               0.98 $\pm$ 0.0141 \\
\bottomrule
\end{tabular}}
\caption{Comparison of Removal based vs Replacement based Anonymization on Semantic Similarity task \label{tab:anon_vs_replacement} of Sec.~\ref{sec:exp_sts_binary}.}
\end{table*}

\begin{table}[t!]
\centering
\scalebox{0.8}{
\begin{tabular}{ll}
\toprule
                           Model Name &    \makecell{Euclidean Distance \\ per perturbation
pair}   \\
\midrule
                    all-mpnet-base-v2 &                         0.642 $\pm$ 0.0016 \\
                 all-distilroberta-v1 &                         0.641 $\pm$ 0.0015 \\
                     all-MiniLM-L6-v2 &                         0.766 $\pm$ 0.0017 \\
                               gemini &                          0.46 $\pm$ 0.0007 \\
           multi-qa-distilbert-cos-v1 &                         0.694 $\pm$ 0.0014 \\
              paraphrase-MiniLM-L6-v2 &                         3.398 $\pm$ 0.0153 \\
 distiluse-base-multilingual-cased-v1 &                          0.638 $\pm$ 0.002 \\
 distiluse-base-multilingual-cased-v2 &                          0.63 $\pm$ 0.0021 \\
paraphrase-multilingual-MiniLM-L12-v2 &                         2.726 $\pm$ 0.0108 \\
            msmarco-distilbert-cos-v5 &                         0.742 $\pm$ 0.0016 \\
           multi-qa-mpnet-base-cos-v1 &                         0.679 $\pm$ 0.0016 \\
               text-embedding-3-small &                          0.67 $\pm$ 0.0013 \\
               text-embedding-3-large &                         0.616 $\pm$ 0.0013 \\
                        voyage-3-lite &                          0.647 $\pm$ 0.001 \\
\bottomrule
\end{tabular}}
\caption{\textbf{Bias Measurement on CMU Book dataset with Euclidean distance as distance function between embeddings}. A distance close to $0$ is better.\label{tab:model_biases_cmu_book_euclidean}}
\end{table}



\end{document}